# Gray-Box Computed Torque Control for Differential-Drive Mobile Robot Tracking

**Arman Javan Sekhavat Pishkhani**  arman.j.sekhavat@ut.ac.ir

University of Tehran, Tehran, Iran

**Abstract**

This study presents a learning-based nonlinear algorithm for tracking control of differential-drive mobile robots. The Computed Torque Method (CTM) suffers from inaccurate knowledge of system parameters, while Deep Reinforcement Learning (DRL) algorithms are known for sample inefficiency and weak stability guarantees. The proposed method replaces the black-box policy network of a DRL agent with a gray-box Computed Torque Controller (CTC) to improve sample efficiency and ensure closed-loop stability. This approach enables finding an optimal set of controller parameters for an arbitrary reward function using only a few short learning episodes. The Twin-Delayed Deep Deterministic Policy Gradient (TD3) algorithm is used for this purpose. Additionally, some controller parameters are constrained to lie within known value ranges, ensuring the RL agent learns physically plausible values. A technique is also applied to enforce a critically damped closed-loop time response. The controller's performance is evaluated on a differential-drive mobile robot simulated in the MuJoCo physics engine and compared against the raw CTC and a conventional kinematic controller.

## 1. Introduction

Mobile robot are employed in a wide variety of settings, including warehouse management, inspection of hazardous environments, and various agricultural tasks [1, 2]. Early mobile robots were mostly teleoperated by human operators [3]; the drive for higher efficiency and autonomy motivated development of automatic control and navigation algorithms. Differential-drive mobile robots (DDMRs) are of particular interest because of low cost, energy efficiency, mechanical simplicity, static stability, and good maneuverability [4]. They consist of two independently driven wheels with a common rotation axis and one or more passive support wheels [5].

Controller design for wheeled mobile robots commonly targets posture regulation and trajectory tracking; trajectory tracking is of greater practical importance because it enables following planned paths to reach goals while avoiding obstacles [5]. A DDMR feedback controller requires the robot's current position and orientation to compare them with desired values. Odometry [5, 6]—often based on incremental rotary encoders—is a common localization method that integrates wheel-motion measurements to estimate posture changes; odometry is frequently fused with IMU, camera, GPS, or LiDAR measurements to reduce drift, often using linear Kalman filters or nonlinear estimators such as EKF/UKF [7].

Over the last decade a wide range of controllers for DDMRs has been proposed [8, 9, 10]. Many trajectory controllers are developed at the kinematic level and therefore neglect dynamic effects such as Coriolis, centrifugal, and dissipative forces; these kinematic schemes commonly use simple linear controllers (e.g., PID) and are often implemented in a hierarchical architecture where a high-level controller commands chassis linear/angular velocities and a low-level loop (typically PID) regulates wheel speeds [11]. While simple and effective at low speeds or for small domains of attraction, such methods degrade when robot dynamics, higher speeds, or tight steady-state requirements matter since neglected dynamic forces become significant [12].

To handle dynamics explicitly, nonlinear and model-based approaches have been studied for DDMR tracking control, including adaptive control [10] and Model Predictive Control (MPC) [8]. Computed Torque Method (CTM) [13] has long been used for robotic manipulators and can be adapted to mobile robots [9] to attain high tracking accuracy. The main limitation of CTM is sensitivity to parametric uncertainty and unmodeled dissipative effects [14]; parameter-identification schemes can mitigate this [15, 16] but add extra complexity to the control design process.

MPC offers constraint handling and optimality [17] but requires substantial online computation, which can be challenging for fast, resource-limited mobile robots. More recently, learning-based methods—particularly Deep Reinforcement Learning (DRL)—have been applied to mobile robot control [18, 19, 20]. They can discover control laws from interaction with the environment and handle complex objectives and constraints. DRL algorithms for control tasks evolved from deep Q-networks [21] to actor-critic approaches including Deep Deterministic Policy Gradient (DDPG) [22] and its modified variant TD3. TD3 is a state-of-the-art algorithm that reduces value overestimation, stabilizes training, and is less sensitive to hyperparameters, compared to DDPG [23]. Model-free DRL can control systems without an explicit dynamic model and find an optimal control law for arbitrary reward functions [24], but major challenges remain: sample inefficiency, weak formal stability guarantees, and potential for exploratory behavior that is damaging to real hardware [25].

Hybrid or gray-box approaches [26] that combine model knowledge with learning have emerged to leverage structure to increase sample efficiency and interpretability. Examples include learning residual terms for model-based controllers [27], using RL to tune PID gains [28], and replacing black-box policies with structured controllers whose parameters are learned [29].

This paper proposes the Gray-box Computed Torque Control (GCTC) algorithm that blends CTC structure with model-free RL for parameter search. Specifically, this study derives a computed torque control law for DDMRs accounting for dissipative (viscous and Coulomb) forces, reformulates and reparametrizes it with a minimal set of unknown parameters to use it in place of the TD3 policy network, improving sample efficiency, stability and interpretability. Some hard inequality constraints are also enforced on trainable policy parameters to ensure physically plausible values are learned.

The proposed GCTC is evaluated on a simulated DDMR in the MuJoCo [30] physics engine with viscous and Coulomb friction included to increase realism. JAX [31] automatic differentiation is used for efficient gradient computations during training. The performance of this controller is compared against a CTC using exact system parameters and also a Lyapunov-based kinematic controller from [5].

## 2. Background

### 2.1. Reinforcement Learning

Markov decision processes (MDPs) are used as a common mathematical formulation to analyze reinforcement learning problems, in which sequential interactions take place between a learning agent and an environment. S denotes the state space, which comprises every possible configuration that the environment can assume. The set of all possible actions that the agent can decide to choose from is denoted by $A$ and is called the action space [24]. For control problems, the plant state vector is taken as the MDP state and the control vector as the action. From a control-systems viewpoint, the agent is considered as the controller while the environment functions as the plant [24]. At each time step during interaction, the environment state $s$ is observed by the learning agent and an action $a$ is chosen by it according to the current state of the environment and its behavior policy $\Pi$. A learning agent's behavior policy falls into two categories: stochastic and deterministic. Under a stochastic policy, the agent randomly selects an action from $A$. On the other hand, deterministic policies map each observed state to a unique action from the action space. In this study only deterministic policies are considered. Under such a policy, the action chosen by the agent is $a = \mu_\Pi(s)$, where $\mu$ is a parametric model with parameter vector $\Pi$. After taking an action, the environment state changes to a new state $s'$ and the agent receives a real number $r$ known as the immediate reward. The reward signal can be considered as a function $r: S \times A \to \mathbb{R}$. Interactions that take place between the agent and the environment result in a sequence of observed states, selected actions and immediate rewards $s_1, a_1, r_1, \ldots, s_n, a_n, r_n$ which is called an episode. The return $r_t^\gamma$ is defined as the total discounted reward that is given to the agent by the environment starting from time step t, $r_t^\gamma = \sum_{k=t}^{+\infty} \gamma^{k-t} r(s_k, a_k)$, where $0 < \gamma < 1$ is the discount factor. Two other important functions are the state value function and the action value function, defined as expected values $V^\mu(s) = \mathbb{E}[r_1^\gamma | s_1 = s; \mu]$ and $Q^\mu(s,a) = \mathbb{E}[r_1^\gamma | s_1 = s, a_1 = a; \mu]$, respectively. A reinforcement learning agent uses trial and error to continuously improve its behavior policy in order to maximize a performance measure, defined by $J(\Pi) = \mathbb{E}[r_1^\gamma | \mu]$. That is, to maximize the expected value of return for all possible state trajectories. This performance measure is a real and scalar-valued function of the policy parameters. The Deterministic Policy Gradient (DPG) theorem [32] makes it possible to estimate the gradient of $J$ with respect to the policy parameters and maximize it by using gradient ascent.

$$\nabla_\Pi J(\Pi) = \mathbb{E}\left[\nabla_a Q_\Gamma^\mu(s,a)|_{a=\mu_\Pi(s)} \frac{\partial}{\partial \Pi} \mu_\Pi(s)\right] \approx N^{-1} \sum \nabla_a Q_\Gamma^\mu(s,a)|_{a=\mu_\Pi(s)} \frac{\partial}{\partial \Pi} \mu_\Pi(s) \quad (1)$$

Where $N$ is the size of a mini-batch of transition tuples $(s, a, r, s')$ used to estimate $\nabla_\Pi J(\Pi)$.

The TD3 employs a neural-network actor $\mu_\Pi$ and uses two other neural-network critics to estimate the corresponding action value function $Q^\mu$. To promote exploration with its deterministic policy, actions are perturbed during training by adding noise. Like DDPG, TD3 uses smooth updates for target policy and Q-networks, to stabilize the training process. However, despite DDPG, TD3 uses delayed policy updates; it updates the target policy network for each $d$ updates of the Q-networks. Additionally, TD3 uses the minimum of two Q-values and adds some clipped noise to actions while computing the Bellman loss [23]. Both Q-networks are then updated to satisfy the Bellman equation,

$$Q^\mu(s,a) = \mathbb{E}[r(s,a) + \gamma Q^\mu(s', \mu(s'))] \tag{2}$$

**2.2. Kinematic Model of DDMR**

The kinematic equations of DDMRs have been previously investigated [5, 33]. A schematic of a DDMR is shown in Fig. 1. To derive the equations of motion for a DDMR, a body frame {B} is rigidly attached to the robot's chassis [33]. The origin of this frame is located at the midpoint of the line segment connecting the centers of the active wheels. The position and orientation of {B} are measured with respect to an inertial frame {U}.

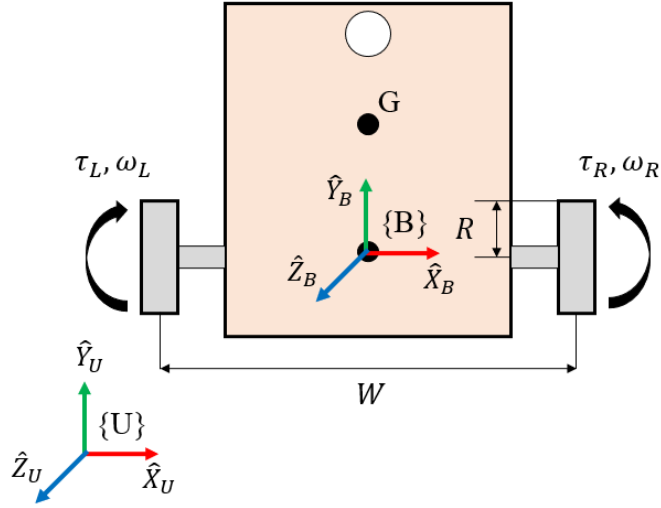

*Figure 1.* Schematic of a DDMR

Let $\boldsymbol{q} = [x, y, \theta, \varphi_R, \varphi_L]^T$ denote the generalized-coordinate vector describing the DDMR configuration, where x and y locate the origin of frame {B}. $\theta$ denotes the DDMR's heading angle measured counterclockwise from the positive x-axis. $\varphi_R$ and $\varphi_L$ denote the rotation angles of the right and left wheels, respectively, measured in radians [33]. Also, let $\omega = \dot{\theta}$, $\omega_R = \dot{\varphi}_R$, $\omega_L = \dot{\varphi}_L$ denote the angular velocities of the chassis, right wheel, and left wheel, respectively. It's convenient to describe the posture of the DDMR with another vector $\boldsymbol{p} := [x, y, \theta]^T$ [33]. The rotation angles of the wheels are also expressed by the vector $\boldsymbol{\Phi} := [\varphi_R, \varphi_L]^T$. Furthermore, two important matrices used in this paper are:

$$\boldsymbol{\Lambda}(\theta) := \frac{1}{R}\begin{bmatrix} \boldsymbol{\Psi} & +\frac{W}{2} \\ \boldsymbol{\Psi} & -\frac{W}{2} \end{bmatrix}_{2\times 3} \tag{3}$$

$$\widetilde{\boldsymbol{\Lambda}}(\theta) := \frac{R}{2}\begin{bmatrix} \boldsymbol{\Psi}^T & \boldsymbol{\Psi}^T \\ +\frac{2}{W} & -\frac{2}{W} \end{bmatrix}_{3\times 2} \tag{4}$$

where

$$\boldsymbol{\Psi}(\theta) := [\cos\theta, \sin\theta] \tag{5}$$

It's easy to see that these matrices satisfy the relation

$$\Lambda(\theta)\tilde{\Lambda}(\theta) = I_2 \quad (6)$$

Based on these notations, the DDMR's nonholonomic motion constraints are given by the following equation [34]:

$$\dot{p} = \tilde{\Lambda}(\theta)\dot{\Phi} \quad (7)$$

By pre-multiplying both sides of (7) by $\Lambda(\theta)$ and using (6) the following important relation is obtained, which will be used later.

$$\Lambda(\theta)\dot{p} = \Lambda(\theta)\tilde{\Lambda}(\theta)\dot{\Phi} = I_2\dot{\Phi} = \dot{\Phi} \quad (8)$$

Additionally, the following relations hold between the linear and angular velocities of the chassis, and the angular velocities of the wheels [34].

$$v = \frac{R}{2}(\omega_R + \omega_L) \quad (9)$$

$$\omega = \frac{R}{W}(\omega_R - \omega_L) \quad (10)$$

**2.3. Dynamic Model of DDMR**

Dynamic equations of motion for the system have been derived in prior studies [12]. From a dynamical-systems perspective, the state of a mechanical system such as DDMR can be represented by a vector comprising the generalized coordinates and their time derivatives [35], namely:

$$x = [x_1, \ldots, x_{10}]^T = [x, \dot{x}, y, \dot{y}, \theta, \omega, \varphi_R, \omega_R, \varphi_L, \omega_L]^T \quad (11)$$

The input vector of this dynamical system is a two-dimensional vector whose elements are the torques applied to the wheels:

$$u = [u_1, u_2]^T = [\tau_R, \tau_L]^T \quad (12)$$

The nonholonomic form of Lagrange's equation [36] can be used to derive the dynamic equations governing the motion of a DDMR.

$$\frac{d}{dt}\left(\frac{\partial T}{\partial \dot{q}_i}\right) - \frac{\partial T}{\partial q_i} = Q_i + \sum_{j=1}^{3}\lambda_j a_{ji} \; ; \; 1 \leq i \leq 5 \quad (13)$$

Here, $T$ is the total kinetic energy of the DDMR, $\lambda_j$ s are the Lagrange multipliers, and $a_{ji}$ s are the coefficients corresponding to the nonholonomic constraints. After simplification and based on the above choice of state variables, the state-space equations of the dynamical system can be expressed as:

$$\dot{x}_1 = x_2 \quad (14)$$
$$\dot{x}_2 = -C_1 x_6 (x_8 + x_{10})\sin x_5 + C_2(u_1 + u_2 + C_3 x_6^2)\cos x_5 \quad (15)$$
$$\dot{x}_3 = x_4 \quad (16)$$
$$\dot{x}_4 = +C_1 x_6 (x_8 + x_{10})\cos x_5 + C_2(u_1 + u_2 + C_3 x_6^2)\sin x_5 \quad (17)$$
$$\dot{x}_5 = x_6 \quad (18)$$
$$\dot{x}_6 = C_4 x_6 (x_8 + x_{10}) + C_5(u_1 - u_2) \quad (19)$$

$$\dot{x}_7 = x_8 \tag{20}$$
$$\dot{x}_8 = C_6 u_1 + C_7 u_2 + C_8 x_6^2 - C_9 x_6 (x_8 + x_{10}) \tag{21}$$
$$\dot{x}_9 = x_{10} \tag{22}$$
$$\dot{x}_{10} = C_7 u_1 + C_6 u_2 + C_8 x_6^2 + C_9 x_6 (x_8 + x_{10}) \tag{23}$$

Where $C_i$ s are constant parameters which depend on both kinematic and dynamic parameters of the DDMR. All of these constants are always positive, except $C_4$ that is always negative and $C_7$ which may be positive or negative.

## 3. Controller Design

### 3.1. Kinematic Controller

The Lyapunov-based kinematic controller used as a benchmark [5] describes the tracking error in a body frame as:

$$\tilde{e} = \begin{bmatrix} \cos\theta & \sin\theta & 0 \\ -\sin\theta & \cos\theta & 0 \\ 0 & 0 & 1 \end{bmatrix} \begin{bmatrix} x_d - x \\ y_d - y \\ \theta_d - \theta \end{bmatrix} \tag{24}$$

Differentiation with respect to time yields:

$$\frac{d}{dt}\tilde{e}_1 = v_d \cos\tilde{e}_3 - v + \tilde{e}_2 \omega \tag{25}$$

$$\frac{d}{dt}\tilde{e}_2 = v_d \sin\tilde{e}_3 - \tilde{e}_1 \omega \tag{26}$$

$$\frac{d}{dt}\tilde{e}_3 = \omega_d - \omega \tag{27}$$

By forming a candidate Lyapunov function it can be shown that the following choice for $v$ and $\omega$ stabilizes the closed-loop system [5]:

$$v = v_d \cos\tilde{e}_3 + k_1 \tilde{e}_1 \tag{28}$$

$$\omega = \omega_d + k_2 v_d \operatorname{sinc}(\tilde{e}_3) \tilde{e}_2 + k_3 \tilde{e}_3 \tag{29}$$

These can then be used to find the appropriate set-point values for the wheel speeds:

$$\omega_{R_d} = \frac{1}{R}\left(v + \frac{W}{2}\omega\right) \tag{30}$$

$$\omega_{L_d} = \frac{1}{R}\left(v - \frac{W}{2}\omega\right) \tag{31}$$

These values are passed to the low-level PID controllers for both wheels, which output the required motor torques.

### 3.2. CTC

Equations (15), (17), and (19) are rewritten for more clarity:

$$\ddot{x} = -C_1\omega(\omega_R + \omega_L)\sin\theta + C_2(\tau_R + \tau_L + C_3\omega^2)\cos\theta \tag{32}$$

$$\ddot{y} = +C_1\omega(\omega_R + \omega_L)\cos\theta + C_2(\tau_R + \tau_L + C_3\omega^2)\sin\theta \tag{33}$$

$$\ddot{\theta} = C_4\omega(\omega_R + \omega_L) + C_5(\tau_R - \tau_L) \tag{34}$$

Multiplying (32) and (33) by $\cos\theta$ and $\sin\theta$, respectively and adding them, yields:

$$\tau_R + \tau_L = -C_3\omega^2 + C_2^{-1}(\ddot{x}\cos\theta + \ddot{y}\sin\theta) \tag{35}$$

Multiplying these equations by $-\sin\theta$ and $\cos\theta$, respectively and addition yields:

$$\omega(\omega_R + \omega_L) = C_1^{-1}(\ddot{y}\cos\theta - \ddot{x}\sin\theta) \tag{36}$$

Substituting (36) in (34) and solving for $\tau_R - \tau_L$ yields:

$$\tau_R - \tau_L = C_5^{-1}\left(\ddot{\theta} - C_4 C_1^{-1}(\ddot{y}\cos\theta - \ddot{x}\sin\theta)\right) \tag{37}$$

For convenience, the following constants are defined:

$$\sigma_1 := \frac{1}{2C_2}, \quad \sigma_2 := -\frac{C_4}{2C_1 C_5}, \quad \sigma_3 := \frac{1}{2C_5}, \quad \sigma_4 := \frac{1}{2}C_3 \tag{38}$$

Therefore, $\sigma_i$ s will always be positive. Also, let:

$$\Delta := \begin{bmatrix} \sigma_1 & -\sigma_2 \\ +\sigma_2 & \sigma_1 \end{bmatrix} \tag{39}$$

$$M(\theta) := \begin{bmatrix} \Psi\Delta^T & +\sigma_3 \\ \Psi\Delta & -\sigma_3 \end{bmatrix} \tag{40}$$

$$C(\omega) := -\sigma_4\omega^2 \begin{bmatrix} 1 \\ 1 \end{bmatrix} \tag{41}$$

After solving (35) and (37) for $\tau_R$ and $\tau_L$, the following compact form is obtained for the equations of motion of the DDMR:

$$u = M(\theta)\ddot{p} + C(\omega) \tag{42}$$

However, this model does not account for dissipative forces. Viscous and Coulomb friction terms are added as follows:

$$u = M(\theta)\ddot{p} + C(\omega) + \tau_V + \tau_D \tag{43}$$

Where $\tau_V$ and $\tau_D$ are the friction-torque vectors for viscous and Coulomb friction, respectively, given by [14]:

$$\tau_V = c_V \dot{\Phi} = c_V \Lambda(\theta)\dot{p} \tag{44}$$

$$\tau_D = c_D \operatorname{sign} \dot{\Phi} = c_D \operatorname{sign} \Lambda(\theta)\dot{p} \tag{45}$$

The controller's objective is to drive the DDMR to follow a prescribed trajectory specified by the desired robot posture:

$$\boldsymbol{p}_d(t) = [x_d(t), y_d(t), \theta_d(t)]^T \tag{46}$$

The tracking error is defined as [5]:

$$\boldsymbol{e}(t) \coloneqq \boldsymbol{p}_d(t) - \boldsymbol{p}(t) \tag{47}$$

The final control law based on the CTM becomes:

$$\boldsymbol{u} = \boldsymbol{M}(\theta)\left(\ddot{\boldsymbol{p}}_d(t) + \boldsymbol{K}_p \boldsymbol{e}(t) + \boldsymbol{K}_i \int_0^t \boldsymbol{e}(t')dt' + \boldsymbol{K}_d \dot{\boldsymbol{e}}(t)\right) + \boldsymbol{C}(\omega) + c_V \Lambda(\theta)\dot{\boldsymbol{p}} + c_D \operatorname{sign} \Lambda(\theta)\dot{\boldsymbol{p}} \tag{48}$$

An integral term of the tracking error is included to reduce steady-state error [35]. Here, $\boldsymbol{K}_p$, $\boldsymbol{K}_i$ and $\boldsymbol{K}_d$ are chosen as diagonal matrices with positive diagonal elements [14]:

$$\boldsymbol{K}_p = \operatorname{diag}(k_p, k_p, k_p') \tag{49}$$

$$\boldsymbol{K}_i = \operatorname{diag}(k_i, k_i, k_i') \tag{50}$$

$$\boldsymbol{K}_d = \operatorname{diag}(k_d, k_d, k_d') \tag{51}$$

Substituting the control law into the dynamic equations of motion, yields the following vector inegro-differential equation in the error space:

$$\ddot{\boldsymbol{e}} + \boldsymbol{K}_p \boldsymbol{e} + \boldsymbol{K}_i \int_0^t \boldsymbol{e}(t')dt' + \boldsymbol{K}_d \dot{\boldsymbol{e}} = 0 \tag{52}$$

Differentiation with respect to time yields:

$$\dddot{\boldsymbol{e}} + \boldsymbol{K}_d \ddot{\boldsymbol{e}} + \boldsymbol{K}_p \dot{\boldsymbol{e}} + \boldsymbol{K}_i \boldsymbol{e} = 0 \tag{53}$$

The corresponding characteristic equation is:

$$\lambda^3 + \boldsymbol{K}_d \lambda^2 + \boldsymbol{K}_p \lambda + \boldsymbol{K}_i = 0 \tag{54}$$

Note that in (54) vector powers are taken elementwise. It's desired for the closed-loop system to be stable and have the fastest non-oscillatory time-response. To derive an appropriate criterion for this, the following lemma is first stated and proved.

**Lemma 1.** Consider an $n$ th order, linear, time-invariant, homogeneous, and stable ODE which has a non-oscillatory time-response. Let $\alpha_1, \ldots \alpha_n$ be the corresponding eigenvalues, which are not all equal. Also, let $\alpha = \min_i \alpha_i$ and $x(t)$ be the time-response of this ODE. Consider another ODE whose eigenvalues are all equal to $\alpha$ and denote its time-response with $x^*(t)$. Then, there exists a time $t'$ such that for all $t \geq t'$, $|x^*(t)| < |x(t)|$.

**Proof:**

Since the $x(t)$ is stable and non-oscillatory, all of its eigenvalues are negative real numbers. The time-responses of these ODEs are:

$$x(t) = F_1(t) \exp \alpha_1 t + \cdots + F_n(t) \exp \alpha_n t \tag{55}$$

$$x^*(t) = (c_1^* + c_2^* t + \cdots + c_n^* t^{n-1}) \exp \alpha t \tag{56}$$

Where $F_i$ s are monomials of the form $c_i t^{k_i}$. Pre-multiplying both sides of these equations by $\exp(-\alpha t)$, yields:

$$\exp(-\alpha t) x(t) = F_1(t) \exp(\alpha_1 - \alpha) t + \cdots + F_n(t) \exp(\alpha_n - \alpha) t \tag{57}$$

and

$$\exp(-\alpha t) x^*(t) = c_1^* + c_2^* t + \cdots + c_n^* t^{n-1} \tag{58}$$

Also let $\tilde{\alpha} = \max_i \alpha_i$. Therefore,

$$|\exp(-\alpha t) x(t)| = |F_1(t) \exp(\alpha_1 - \tilde{\alpha}) t + \cdots + F_n(t) \exp(\alpha_n - \tilde{\alpha}) t| \times \exp(\tilde{\alpha} - \alpha) t =$$

$$|F(t)| \exp(\tilde{\alpha} - \alpha) t \tag{59}$$

$F(t)$ is a summation of some monomials and some exponentially-decayed monomials. Therefore, after some time, $|F(t)| \geq M_1$ holds for some $M_1 > 0$. Hence:

$$|\exp(-\alpha t) x(t)| \geq M_1 \exp(\tilde{\alpha} - \alpha) t \tag{60}$$

Additionally, applying the triangle inequality yields the following for $t > 1$:

$$|\exp(-\alpha t) x^*(t)| = |c_1^* + \cdots + c_n^* t^{n-1}| \leq |c_1^*| + \cdots + |c_n^*| t^{n-1} \leq (|c_1^*| + \cdots + |c_n^*|) t^{n-1} = M_2 t^{n-1} \tag{61}$$

Using the L'Hôpital's rule,

$$\lim_{t \to +\infty} \frac{\exp(\tilde{\alpha} - \alpha) t}{t^{n-1}} = +\infty \tag{62}$$

Therefore, after some time, the inequality

$$\frac{\exp(\tilde{\alpha} - \alpha) t}{t^{n-1}} > \frac{M_2}{M_1} \tag{63}$$

holds. Finally,

$$|\exp(-\alpha t) x(t)| \geq M_1 \exp(\tilde{\alpha} - \alpha) t > M_2 t^{n-1} \geq |\exp(-\alpha t) x^*(t)| \tag{64}$$

Hence $|x^*(t)| < |x(t)|$. This completes the proof. ∎

This lemma shows that to achieve the fastest non-oscillatory time-response, the poles of the closed-loop system must be equal negative numbers. To enforce the satisfaction of these criteria, the following characteristic polynomial is chosen:

$$P(\lambda) = (\lambda + [\alpha^2 + \varepsilon, \alpha^2 + \varepsilon, \beta^2 + \varepsilon]^T)^3 \tag{65}$$

After expansion, the controller parameters become:

$$k_p = 3(\alpha^2 + \varepsilon)^2, \; k_i = (\alpha^2 + \varepsilon)^3, \; k_d = 3(\alpha^2 + \varepsilon) \tag{66}$$

and

$$k'_p = 3(\beta^2 + \varepsilon)^2, \; k'_i = (\beta^2 + \varepsilon)^3, \; k'_d = 3(\beta^2 + \varepsilon) \tag{67}$$

Where $\alpha$ and $\beta$ are real-valued free parameters. $\varepsilon$ is a positive, non-trainable scalar chosen arbitrarily to ensure the closed-loop system poles are strictly negative.

### 3.3. GCTC

The state observed by an RL agent must include all of the information required for decision making [37]. Hence, a suitable choice for the state vector is:

$$s = \begin{bmatrix} \boldsymbol{e} \\ \int_0^t \boldsymbol{e}(t')dt' \\ \dot{\boldsymbol{e}} \\ [\dot{x}, \dot{y}, \omega]^T \\ \ddot{\boldsymbol{p}}_d \\ \theta \end{bmatrix} \tag{68}$$

For each state-action pair, the following reward is given to the agent:
$$r(s, a) = \text{sech}(\boldsymbol{E}^T \boldsymbol{H}_e \boldsymbol{E} + \boldsymbol{u}^T \boldsymbol{H}_u \boldsymbol{u}) \tag{69}$$

Where $\boldsymbol{E} := \left[\boldsymbol{e}^T \;\middle|\; \int_0^t \boldsymbol{e}^T(t')dt' \;\middle|\; \dot{\boldsymbol{e}}^T\right]^T$, $\boldsymbol{H}_e$ is a positive semi-definite matrix and $\boldsymbol{H}_u$ is a positive definite matrix. The term $\boldsymbol{E}^T \boldsymbol{H}_e \boldsymbol{E}$ penalizes tracking error and the term $\boldsymbol{u}^T \boldsymbol{H}_u \boldsymbol{u}$ penalizes energy consumption. This choice for the reward signal encourages the agent to learn behavior policies that minimize the tracking error, while avoiding excessively large control efforts.

$\sigma_1, \sigma_2, \sigma_3, \sigma_4, c_V, c_D, \alpha,$ and $\beta$ are trainable. During the policy optimization, it's possible for the agent to learn incorrect and physically implausible values for $\sigma_1, \sigma_2, \sigma_3, \sigma_4, c_V, c_D$. To address this issue, these values are constrained to lie within predefined (guessed) ranges. This leads to a constrained optimization problem. The constrained problem is then transformed into an unconstrained one using the following change of variables [38]:

$$\sigma_i = \sigma_{c_i} + \sigma_{r_i} \tanh z_i \; ; \; 1 \leq i \leq 4 \tag{70}$$

$$c_V = c_{V_c} + c_{V_r} \tanh z_V \tag{71}$$

$$c_D = c_{D_c} + c_{D_r} \tanh z_D \tag{72}$$

Where $z_i$ s, $z_V$ and $z_D$ are free trainable parameters. $\sigma_{c_i}$ s, $c_{V_c}$ and $c_{D_c}$ are the centers of the specified ranges and $\sigma_{r_i}$ s, $c_{V_r}$ and $c_{D_r}$ are their radii. Therefore, the policy parameters become:

$$\Pi = [z_1, z_2, z_3, z_4, z_V, z_D, \alpha, \beta]^T \tag{73}$$

If the tracking error exceeds a threshold during an episode, the episode is terminated. Additionally, to prevent motor torques that cause slippage or tip-over [39], the actor's output is clipped.

**Algorithm 1** GCTC

---

Randomly initialize critic networks $Q_{\Gamma_1}, Q_{\Gamma_2}$ and policy parameters $z_i$ s, $z_V, z_D, \alpha, \beta$

Initialize target parameters $\Gamma'_1 \leftarrow \Gamma_1,\ \Gamma'_2 \leftarrow \Gamma_2,\ \Pi' \leftarrow \Pi$

Initialize experience replay buffer B

**while** $\|e(t)\| <$ max_track_error **and** $t <$ max_episode_len **do**

    Compute $\sigma_i$ s, $c_V, c_D, \boldsymbol{K_p, K_i, K_d}$

    Select action with exploration noise $a = \mu_\Pi(s) + \epsilon,\ \epsilon \sim N(0, \tilde{\sigma})$

    Apply the clipped control signal to the plant $u = \text{clip}(a, -\tau_{\max}, +\tau_{\max})$

    Observe the transition tuple $(s, a, r, s')$ and store it in B

    Randomly sample a mini-batch of size $N$ from B

    $\tilde{a} \leftarrow \mu_{\Pi'}(s') + \epsilon',\ \epsilon' \sim \text{clip}(N(0, \tilde{\sigma}'), -c, +c)$

    $y \leftarrow r + \gamma \min_{i=1,2} Q_{\Gamma'_i}(s', \tilde{a})$

    Update critics by gradient descent: $\Gamma_i \leftarrow \underset{\Gamma_i}{\text{argmin}}\ N^{-1} \sum \left(y - Q_{\Gamma_i}(s, a)\right)^2\ ;\ i = 1, 2$

    **if** t **mod** d **then**

        Update $\Pi$ by the DPG theorem and gradient ascent:

$$\nabla_\Pi J(\Pi) \approx N^{-1} \sum \nabla_a Q_{\Gamma_1}(s, a)|_{a=\mu_\Pi(s)} \frac{\partial}{\partial \Pi} \mu_\Pi(s)$$

        Update target networks:

$$\Gamma'_i \leftarrow \eta \Gamma_i + (1 - \eta)\Gamma'_i\ ;\ i = 1, 2$$

$$\Pi' \leftarrow \eta \Pi + (1 - \eta)\Pi'$$

    **end if**

**end while**

---

## 4. Experiments

A single sinusoidal path was used for training. Only 11 episodes, each with a maximum duration of 5 seconds, were used. The learned control law was evaluated on three test trajectories: a different high-velocity sinusoid, a circle with time-varying speed, and a square. The same experiments were performed with the kinematic controller and a CTC using exact parameters; the latter represents an upper bound on tracking accuracy.

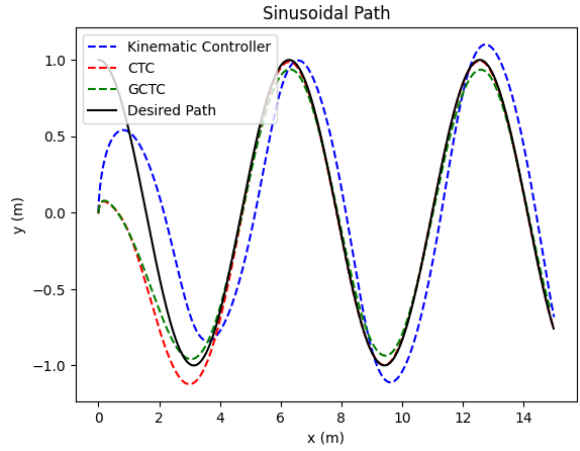

*Figure 2.* Trajectories generated by different controllers for a sinusoidal desired path

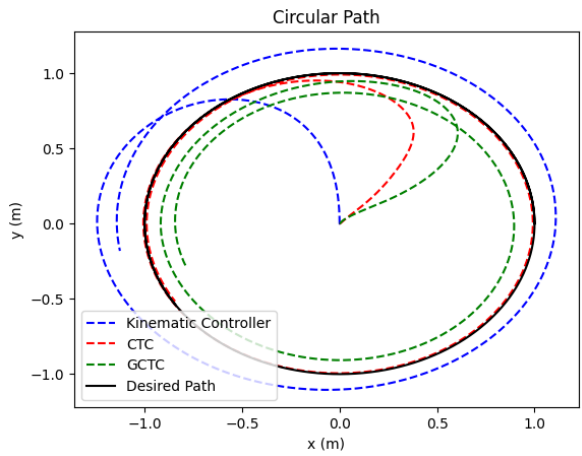

*Figure 3.* Trajectories generated by different controllers for a circular desired path

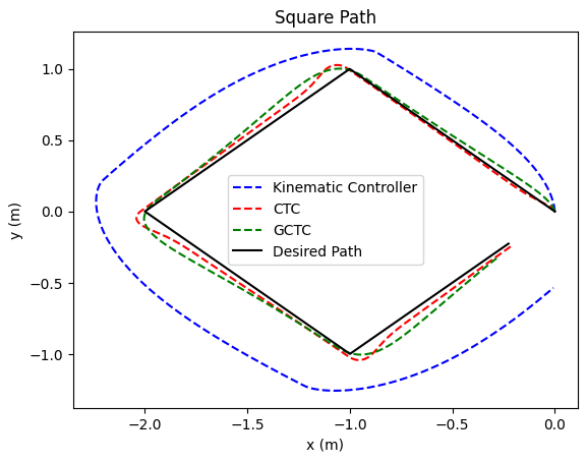

*Figure 4.* Trajectories generated by different controllers for a square desired path

## 5. Conclusion

Experiments indicate that the proposed algorithm can concurrently identify the system and design the controller for DDMRs. By replacing the policy network with a gray-box model, the algorithm requires only a few short training episodes to converge to an optimal solution. Furthermore, the added constraints guarantee a stable and critically damped closed-loop time response. The closed-loop system also benefits from a nonlinear controller, yielding a larger domain of attraction, smaller steady-state errors, and improved tracking of high-velocity trajectories. After training, the resulting controller can be implemented on a low-cost microcontroller for real-time applications because of its low computational complexity. All of this is achievable without accurate prior knowledge of the dynamic parameters.